\pdfoutput=1  
\documentclass[11pt]{article}

\usepackage[T1]{fontenc}
\usepackage[utf8]{inputenc}
\usepackage{amsmath}          
\usepackage{newtxtext,newtxmath}
\usepackage{amssymb}
\usepackage[protrusion=true,expansion=true]{microtype}
\usepackage[letterpaper,margin=1.15in]{geometry}
\usepackage{booktabs}
\usepackage{threeparttable}
\usepackage{graphicx}
\usepackage{subcaption}
\usepackage{xcolor}
\usepackage{enumitem}
\usepackage{algorithm}
\usepackage{algpseudocode}
\usepackage{tikz}
\usetikzlibrary{arrows.meta,positioning,calc}
\usepackage[round,authoryear]{natbib}
\usepackage[colorlinks=true,allcolors=RoyalBlue,breaklinks=true]{hyperref}
\definecolor{RoyalBlue}{HTML}{2a78d6}
\definecolor{inkmid}{HTML}{52514E}
\definecolor{coral}{HTML}{EB6834}  \definecolor{coralbg}{HTML}{FCEBE4}  \definecolor{coralln}{HTML}{F0B79C}
\definecolor{mint}{HTML}{1BAF7A}   \definecolor{mintbg}{HTML}{E3F6EF}   \definecolor{mintln}{HTML}{97DCC3}
\definecolor{blueln}{HTML}{A9C6EA} \definecolor{bluebg}{HTML}{E4EDF9}
\definecolor{greybg}{HTML}{F0F0EE} \definecolor{greyln}{HTML}{C9C9C4}

\graphicspath{{figures/}}
\usepackage[font=small,labelfont=bf,skip=6pt]{caption}
\setlist[itemize]{topsep=3pt,itemsep=2pt,parsep=0pt,leftmargin=1.4em}
\setlist[enumerate]{topsep=3pt,itemsep=2pt,parsep=0pt,leftmargin=1.6em}

\newcommand{\orc}{\textsc{Orcetra}}
\newcommand{\flaml}{\textsc{FLAML}}
\newcommand{\ag}{\textsc{AutoGluon}}

\newcommand{\AgEnsemblePct}{100}
\newcommand{\BtBeat}{6.7}
\newcommand{\BtBrier}{0.1065}
\newcommand{\BtDays}{7}
\newcommand{\BtMarketBrier}{0.0936}
\newcommand{\BtN}{15}

\newcommand{\CalBelowBins}{6}
\newcommand{\CalBelowExp}{60}
\newcommand{\CalBelowN}{160}
\newcommand{\CalBelowObs}{50}
\newcommand{\CalBelowP}{0.087}
\newcommand{\CalBelowZ}{1.7}
\newcommand{\CalN}{178}
\newcommand{\CalNBins}{10}
\newcommand{\CalWorstActual}{0.12}

\newcommand{\CalWorstMarket}{0.25}
\newcommand{\CalWorstN}{33}
\newcommand{\CalWorstSE}{0.075}
\newcommand{\CalWorstZ}{1.7}
\newcommand{\CalWrongSign}{5}

\newcommand{\DupNames}{67}
\newcommand{\FairAg}{27.3}
\newcommand{\FairFl}{28.0}
\newcommand{\FairHOrcAgL}{62}
\newcommand{\FairHOrcAgP}{6.6\times 10^{-1}}
\newcommand{\FairHOrcAgW}{68}
\newcommand{\FairHOrcFlL}{62}
\newcommand{\FairHOrcFlP}{3.9\times 10^{-1}}
\newcommand{\FairHOrcFlW}{73}
\newcommand{\FairKilledN}{0}
\newcommand{\FairKilledPct}{0}
\newcommand{\FairMatchedAg}{27.9}
\newcommand{\FairMatchedFl}{31.7}
\newcommand{\FairMatchedOrc}{32.7}
\newcommand{\FairMatchedTie}{7.7}
\newcommand{\FairN}{143}
\newcommand{\FairNCls}{105}
\newcommand{\FairNEvalMax}{564}
\newcommand{\FairNEvalMed}{16}
\newcommand{\FairNReg}{38}

\newcommand{\FairOrc}{34.3}
\newcommand{\FairOrigAg}{19.6}
\newcommand{\FairOrigFl}{11.2}

\newcommand{\FairOrigOrc}{59.4}
\newcommand{\FairOrigTie}{9.8}
\newcommand{\FairPairedHonestAg}{21.2}
\newcommand{\FairPairedHonestFl}{29.8}
\newcommand{\FairPairedHonestOrc}{40.4}
\newcommand{\FairPairedHonestTie}{8.7}
\newcommand{\FairPairedOracleAg}{26.0}
\newcommand{\FairPairedOracleFl}{27.9}
\newcommand{\FairPairedOracleOrc}{37.5}
\newcommand{\FairPairedOracleTie}{8.7}

\newcommand{\FairSearchMax}{185}
\newcommand{\FairSearchMed}{68}
\newcommand{\FairSearchOverPct}{100}
\newcommand{\FairTie}{10.5}
\newcommand{\FairTimeOrcMed}{72}
\newcommand{\FeatMax}{477}
\newcommand{\FeatMed}{16}
\newcommand{\FeatMin}{1}
\newcommand{\GapClsMaxPts}{3.25}
\newcommand{\GapClsMeanPts}{0.27}

\newcommand{\GapClsQthreePts}{0.11}

\newcommand{\HAgFlL}{147}

\newcommand{\HAgFlT}{28}

\newcommand{\HAgFlW}{338}
\newcommand{\HAgFlWp}{65.9}
\newcommand{\HOrcAgL}{135}
\newcommand{\HOrcAgLp}{26.3}
\newcommand{\HOrcAgP}{1.9\times 10^{-21}}
\newcommand{\HOrcAgT}{38}

\newcommand{\HOrcAgW}{340}
\newcommand{\HOrcAgWp}{66.3}
\newcommand{\HOrcFlL}{91}
\newcommand{\HOrcFlLp}{17.7}
\newcommand{\HOrcFlP}{9.5\times 10^{-46}}
\newcommand{\HOrcFlT}{29}

\newcommand{\HOrcFlW}{393}
\newcommand{\HOrcFlWp}{76.6}
\newcommand{\InflPredMed}{1.30}
\newcommand{\InflPredQone}{0.78}
\newcommand{\InflPredQthree}{1.65}

\newcommand{\KSel}{40}
\newcommand{\KSelHi}{60}
\newcommand{\KSelLo}{20}
\newcommand{\MarginCloseP}{78}
\newcommand{\MarginMed}{+0.10}
\newcommand{\MatchAgAg}{29.2}
\newcommand{\MatchAgFl}{20.0}
\newcommand{\MatchAgN}{65}
\newcommand{\MatchAgOrc}{43.1}
\newcommand{\MatchAgTie}{7.7}
\newcommand{\MatchBothAg}{27.0}
\newcommand{\MatchBothFl}{20.6}
\newcommand{\MatchBothN}{63}
\newcommand{\MatchBothOrc}{44.4}
\newcommand{\MatchBothTie}{7.9}
\newcommand{\MatchFlAg}{23.0}
\newcommand{\MatchFlFl}{16.1}
\newcommand{\MatchFlN}{174}
\newcommand{\MatchFlOrc}{49.4}
\newcommand{\MatchFlTie}{11.5}
\newcommand{\MeanAccAg}{0.7320}
\newcommand{\MeanAccBar}{0.749}
\newcommand{\MeanAccFl}{0.7179}
\newcommand{\MeanAccOrc}{0.7494}
\newcommand{\MergedAg}{18.5}
\newcommand{\MergedFl}{9.9}
\newcommand{\MergedN}{513}
\newcommand{\MergedOrc}{61.2}
\newcommand{\MergedTie}{10.3}
\newcommand{\NCls}{382}
\newcommand{\NData}{513}

\newcommand{\NReg}{131}
\newcommand{\NRegRel}{127}
\newcommand{\OverBudgetPct}{78}
\newcommand{\OverBudgetTwoX}{50}
\newcommand{\PairedN}{104}
\newcommand{\PairedNCls}{79}
\newcommand{\RankAg}{2.01}
\newcommand{\RankFl}{2.48}
\newcommand{\RankOrc}{1.51}
\newcommand{\RatioOrcAgMed}{2.24}
\newcommand{\RatioOrcFlMed}{1.46}
\newcommand{\RelMseAg}{1.02}
\newcommand{\RelMseFl}{1.20}
\newcommand{\RelMseGeoAg}{1.74}
\newcommand{\RelMseGeoFl}{2.20}
\newcommand{\RelMseGeoOrc}{1.09}
\newcommand{\RelMseOrc}{1.00}
\newcommand{\SampMax}{100,000}
\newcommand{\SampMed}{27,901}
\newcommand{\SampMin}{240}
\newcommand{\SelRulePts}{4.8}
\newcommand{\SigmaMedPts}{0.56}
\newcommand{\SpliceAgWorse}{88}
\newcommand{\SpliceBetter}{3}
\newcommand{\SpliceEnsembleN}{0}
\newcommand{\SpliceFlWorse}{59}
\newcommand{\SpliceN}{131}
\newcommand{\SpliceNewAg}{28}

\newcommand{\SpliceNewOrc}{91}

\newcommand{\SpliceOldAg}{44}

\newcommand{\SpliceOldOrc}{70}

\newcommand{\SpliceSame}{117}
\newcommand{\SpliceTimeNew}{146}
\newcommand{\SpliceTimeOld}{97}
\newcommand{\SpliceWorse}{11}
\newcommand{\SqrtTwoLnK}{2.72}
\newcommand{\SqrtTwoLnKHi}{2.86}
\newcommand{\SqrtTwoLnKLo}{2.45}
\newcommand{\SweepImpMed}{3.6}
\newcommand{\SweepIterMed}{48}
\newcommand{\SweepN}{1,635}
\newcommand{\SweepNCls}{1357}
\newcommand{\SweepNReg}{278}
\newcommand{\SweepTimeMax}{4,653}
\newcommand{\SweepTimeMed}{38}
\newcommand{\SweepWinPct}{87.8}
\newcommand{\TestSizeMed}{6,000}

\newcommand{\TimeAgMed}{62}

\newcommand{\TimeFlMax}{689}

\newcommand{\TimeFlMed}{70}

\newcommand{\TimeOrcMax}{1351}
\newcommand{\TimeOrcMean}{146}
\newcommand{\TimeOrcMed}{120}
\newcommand{\TimeOrcPninety}{219}
\newcommand{\TopModelFl}{lgbm}
\newcommand{\TopModelFlPct}{47.8}
\newcommand{\TopModelOrc}{HistGradientBoosting}
\newcommand{\TopModelOrcN}{132}

\newcommand{\TraceInflKten}{0.13}

\newcommand{\TraceInflKtwo}{0.11}

\newcommand{\TraceInflMax}{0.27}
\newcommand{\TraceKMax}{80}
\newcommand{\TraceN}{58}

\newcommand{\TwoDedup}{540}
\newcommand{\TwoErr}{27}
\newcommand{\TwoFl}{14.6}
\newcommand{\TwoN}{513}
\newcommand{\TwoOrc}{78.4}
\newcommand{\TwoRatioMed}{0.99}
\newcommand{\TwoTie}{7.0}
\newcommand{\TwoTimeFlMed}{31.3}
\newcommand{\TwoTimeOrcMed}{30.0}
\newcommand{\UniqAg}{19.0}
\newcommand{\UniqFl}{12.2}
\newcommand{\UniqHOrcAgL}{90}
\newcommand{\UniqHOrcAgP}{2.1\times 10^{-19}}
\newcommand{\UniqHOrcAgW}{255}
\newcommand{\UniqHOrcFlL}{69}
\newcommand{\UniqHOrcFlP}{1.0\times 10^{-32}}
\newcommand{\UniqHOrcFlW}{287}
\newcommand{\UniqN}{378}
\newcommand{\UniqNCls}{293}
\newcommand{\UniqNReg}{85}
\newcommand{\UniqOrc}{57.4}
\newcommand{\UniqTie}{11.4}
\newcommand{\WinAgAll}{21.6}
\newcommand{\WinAgCls}{17.5}
\newcommand{\WinAgReg}{33.6}
\newcommand{\WinFlAll}{10.9}
\newcommand{\WinFlCls}{11.3}
\newcommand{\WinFlReg}{9.9}
\newcommand{\WinMarginCloseP}{63}
\newcommand{\WinMarginMed}{0.77}
\newcommand{\WinOrcAll}{57.1}
\newcommand{\WinOrcCls}{58.4}
\newcommand{\WinOrcReg}{53.4}
\newcommand{\WinTieAll}{10.3}
\newcommand{\WinTieCls}{12.8}
\newcommand{\WinTieReg}{3.1}
\newcommand{\ZsBeat}{37.9}
\newcommand{\ZsBrier}{0.1463}
\newcommand{\ZsMarketBrier}{0.1468}
\newcommand{\ZsN}{29}

\newcommand{\figorpending}[2]{%
  \IfFileExists{figures/#1}{\includegraphics[width=#2]{#1}}%
  {\fbox{\parbox[c][3.2cm][c]{\dimexpr#2-2\fboxsep-2\fboxrule\relax}%
   {\centering\small\itshape figure pending:\\ protocol-corrected run still in progress}}}}

\newcommand{\logowidth}{0.085}
\newcommand{\archfigure}{%
  \begin{tikzpicture}
    \node[anchor=north west,inner sep=0] (diag) at (0,0)
      {\resizebox{\linewidth}{!}{
\begin{tikzpicture}[
    font=\sffamily\small,
    box/.style   = {rounded corners=2.5pt, draw, line width=0.5pt,
                    minimum width=2.7cm, minimum height=1.0cm,
                    align=center, inner sep=2pt},
    flow/.style  = {-{Latex[length=1.7mm,width=1.5mm]}, line width=0.5pt, draw=inkmid},
    hop/.style   = {-{Latex[length=1.7mm,width=1.5mm]}, line width=0.7pt},
    lane/.style  = {font=\sffamily\footnotesize, text=inkmid, anchor=east},
    tag/.style   = {font=\itshape\scriptsize, anchor=center},
  ]

  \node[box, fill=coralbg, draw=coralln] (a1) at (0,0)    {Tables};
  \node[box, fill=coralbg, draw=coralln] (a2) at (3.5,0)  {Train\,$\vert$\,Test};
  \node[box, fill=coralbg, draw=coralln] (a3) at (7.0,0)  {Search};
  \node[box, fill=coralbg, draw=coralln] (a4) at (10.5,0) {Best on test};
  \node[box, fill=greybg,  draw=greyln]  (a5) at (14.0,0) {Win rate};
  \foreach \i/\j in {a1/a2, a2/a3, a3/a4, a4/a5} \draw[flow] (\i) -- (\j);

  \draw[hop, draw=coral, dashed, dash pattern=on 2.2pt off 1.6pt]
        (a2.north) .. controls +(0,1.0) and +(0,1.0) .. (a3.north);
  \node[tag, text=coral] at (5.25,1.28) {reads the test split};

  \draw[coral, line width=0.6pt, dash pattern=on 1.6pt off 1.6pt]
        ([xshift=-3pt,yshift=-3pt]a3.south west) rectangle
        ([xshift=3pt,yshift=3pt]a3.north east);

  \node[lane] at (-1.75,0) {original};

  \begin{scope}[yshift=-3.7cm]
    \node[box, fill=mintbg, draw=mintln] (b1) at (0,0)    {Tables};
    \node[box, fill=mintbg, draw=mintln] (b2) at (3.5,0)  {Train\,$\vert$\,Val\,$\vert$\,Test};
    \node[box, fill=mintbg, draw=mintln] (b3) at (7.0,0)  {Search};
    \node[box, fill=mintbg, draw=mintln] (b4) at (10.5,0) {Best on val, refit};
    \node[box, fill=bluebg, draw=blueln] (b5) at (14.0,0) {Win rate};
    \foreach \i/\j in {b1/b2, b2/b3, b3/b4, b4/b5} \draw[flow] (\i) -- (\j);

    \draw[hop, draw=mint] (b2.north) .. controls +(0,1.0) and +(0,1.0) .. (b3.north);
    \node[tag, text=mint] at (5.25,1.28) {reads validation only};

    \draw[mint, line width=0.8pt] ([xshift=-2pt,yshift=-2pt]b3.south west)
          rectangle ([xshift=2pt,yshift=2pt]b3.north east);

    \node[lane] at (-1.75,0) {corrected};
  \end{scope}

  \node[tag, text=coral, anchor=north] at (7.0,-1.32) {deadline is advisory};
  \node[tag, text=mint,  anchor=north] at (7.0,-5.02) {deadline is enforced};

\end{tikzpicture}
}};
    \IfFileExists{figures/logo.png}{%
      \node[anchor=north west,inner sep=0] at ([yshift=1.05\baselineskip]diag.north west)
        {\includegraphics[width=\logowidth\linewidth]{logo.png}};}{}
  \end{tikzpicture}}

\title{\bfseries Winning by Peeking:\\[2pt]
\large Unenforced Budgets and Test-Set Selection Inflate\\
Short-Budget AutoML Comparisons}

\author{%
  Guilin Zhang\\
  \small Independent Researcher\\
  \small\texttt{guilindev@gmail.com}
  \and
  Kai Zhao\\
  \small Independent Researcher
}
\date{\today}

\begin{document}
\maketitle

\begin{abstract}
\noindent
Comparisons between AutoML systems at short time budgets --- tens of seconds
rather than hours --- are common in tool READMEs and workshop papers, and they
are easy to get wrong. We report a case study in which a simple AutoML engine,
\orc{}, appeared to beat \flaml{} and \ag{} on \NData{} OpenML datasets, winning
\WinOrcAll{}\% of them at a nominal 60-second budget and \TwoOrc{}\% of
of datasets against \flaml{} alone at 30 seconds. Both margins came from
protocol defects that a results table cannot show. The search loop scored every
candidate model on the test split and reported the best, making the headline
metric a maximum over dozens of noisy estimates while the baselines selected on
training data and touched the test set once; and the budget was checked before
launching a candidate but never enforced during one, so the system consumed a
median of \TimeOrcMed{}\,s against a 60-second budget ---
\RatioOrcAgMed{}$\times$ the wall-clock \ag{} used. Re-running with selection
moved to a validation split, the deadline enforced externally and every framework
pinned to an equal share of the machine, \orc{}'s win rate on the re-run subset
falls from \FairOrigOrc{}\% to \FairOrc{}\% and no pairwise difference against
either competitor remains significant. Recording both estimands inside a single
search lets us attribute the collapse: the selection rule accounts for \SelRulePts{} percentage points and
unequal compute for most of the rest. The same traces give the selection bias as
a function of budget, measured rather than assumed: it grows with $K$ but reaches
only \TraceInflMax{} accuracy points, about five times smaller than the
$\sigma\sqrt{2\ln K}$ bound that a marginal-standard-error argument predicts,
because candidates scored on shared test rows cancel most of the noise. We close
with a checklist for short-budget comparisons. Along the way we describe \orc{} itself, a 1{,}661-line
search loop over a fixed scikit-learn model pool, and the prediction-market
calibration study that motivated it. Code, per-dataset results and the scripts
that regenerate every number and figure are released with the paper.
\end{abstract}

\section{Introduction}
\label{sec:intro}

The published AutoML benchmarks that practitioners trust are run at budgets of
one to four hours per dataset \citep{gijsbers2024amlb}. The comparisons that
practitioners actually read are not. A tool's README, a blog post, or a
workshop submission will report that a new system beats \flaml{} or \ag{} on
some collection of datasets at 30 or 60 seconds per task, because that is the
regime in which an interactive tool is used and it is also the only regime in
which a solo developer can afford to sweep several hundred datasets. The
short-budget regime is legitimate and worth studying. It is also, we argue, where
protocol errors do the most damage and are least likely to be caught.

This paper is a case study in exactly that failure, conducted on our own system.

\orc{} is a small AutoML engine we wrote: it detects the task type, runs a fixed
pool of scikit-learn baselines, then spends the remaining budget on guided random
search over model families and hyperparameters, with a deduplication cache and a
post-hoc weighted ensemble of the best few models. It is 1{,}661 lines of
Python and installs with \texttt{pip} in seconds. Benchmarked against \flaml{}
\citep{wang2021flaml} and \ag{} \citep{erickson2020autogluon} on \NData{} OpenML
datasets at a nominal 60-second budget, it won \WinOrcAll{}\% of datasets against
\WinAgAll{}\% for \ag{} and \WinFlAll{}\% for \flaml{}. Against \flaml{} alone at
30 seconds it won \TwoOrc{}\%. Both margins are large, stable across
classification and regression, and significant at $p = \HOrcFlP{}$ and
$p = \HOrcAgP{}$ respectively. They are also, in large part, wrong.

Three defects produced them. None is visible in a results table, and we found
the third only while auditing the files for this paper.

\paragraph{Selection on the test set.} \orc{}'s search loop scored each candidate
pipeline by fitting it on the training split and evaluating it on the
\emph{test} split, then reported the best score it had seen. Over a 60-second
budget the loop evaluates a median of several dozen candidates, so the reported
number is a maximum over dozens of noisy estimates of the same quantity, each
computed on the same held-out sample. \flaml{} and \ag{} select internally by
cross-validation on training data and touch the test set exactly once. The
comparison therefore contrasts an oracle-selected score against two honestly
held-out scores. This is the classical selection bias in performance estimation
\citep{cawley2010overfitting}, with one twist that matters here: the number of
selection events grows with the compute budget, so the bias grows with the very
knob the experiment is varying.

\paragraph{Budgets checked but not enforced.} The loop tested
\texttt{elapsed < budget} before starting a candidate and never interrupted one
in flight. A random-forest fit launched at $t=59$\,s on a 100k-row dataset runs
to completion regardless. \flaml{} and \ag{} bound their overshoot ---
\ag{} tightly, \flaml{} loosely --- by doing fewer trials under contention. The
consequence is visible in the logs and nowhere in the results: at a nominal
60-second budget, \orc{} consumed a median of \TimeOrcMed{}\,s per dataset
against \TimeFlMed{}\,s for \flaml{} and \TimeAgMed{}\,s for \ag{}, exceeding
its own budget on \OverBudgetPct{}\% of datasets and doubling it on
\OverBudgetTwoX{}\%.

\paragraph{Results waiting to be spliced.} A later regression-only re-run sits in
the result directory beside the original sweep, with nothing marking which
supersedes which. Deduplicating every result file by dataset ID --- the obvious
move, and the one our own first analysis made --- merges them and lifts the
headline to \MergedOrc{}\%. The re-run was supposed to measure a new ensembling
step; the ensemble never wins, and \orc{}'s own scores are unchanged on
\SpliceSame{} of \SpliceN{} datasets. What moved was the competitors, which
scored worse under the heavier machine load of the later run
(\S\ref{sec:splice}).

We measured all three rather than disclosing them. Our contributions:

\begin{enumerate}
\item \textbf{The original result, reported in full} (\S\ref{sec:naive}), with
per-task-type breakdowns, head-to-head counts, mean ranks and wall-clock
distributions, so that the corrected numbers can be compared against something
concrete rather than a summary.

\item \textbf{A measurement of each confound in isolation}
(\S\ref{sec:confounds}). Restricting to the datasets on which \orc{} consumed no
more wall-clock than \flaml{} drops its win rate from \WinOrcAll{}\% to
\MatchFlOrc{}\%; restricting to those where it also used no more than \ag{}
drops it to \MatchBothOrc{}\%. We also give a closed-form estimate of the
selection bias from two numbers any harness logs, and later check it against
measurement.

\item \textbf{A protocol-corrected re-run} (\S\ref{sec:corrected}) in which
model selection moves to a validation split, the test set is evaluated once, the
deadline is enforced from outside the search, and every framework gets an equal
pinned share of the machine. \orc{}'s win rate on those datasets falls from
\FairOrigOrc{}\% to \FairOrc{}\%, and no pairwise comparison against \flaml{}
or \ag{} remains significant. Because the same run records what the old protocol
would have reported, we can attribute the collapse: \SelRulePts{} points to the
selection rule, most of the rest to compute.

\item \textbf{The selection bias as a measured function of budget}
(\S\ref{sec:curve}). Logging every candidate's test score turns the
order-statistics argument into a measurement. The bias does grow with $K$, but
reaches only \TraceInflMax{} accuracy points --- about five times below the
$\sigma\sqrt{2\ln K}$ bound --- because candidates share test rows and the common
noise cancels. We report the bound as a screen and its overestimate as a
calibration of that screen.

\item \textbf{A checklist} (\S\ref{sec:reco}) for short-budget AutoML
comparisons, derived from what went wrong here rather than from first
principles.

\item \textbf{The system and a calibration case study}
(\S\ref{sec:system}, \S\ref{sec:calibration}). \orc{} began as a forecasting
tool for prediction markets. On \CalN{} resolved Polymarket contracts the
framework recovers something that leans towards the textbook
favourite--longshot bias but does not reach significance once the sample is
pooled rather than mined bin by bin --- an instance of the paper's own thesis in a
second domain --- and its language-model forecaster lost to market consensus on
the small backtests we ran.
\end{enumerate}

\begin{figure}[t]
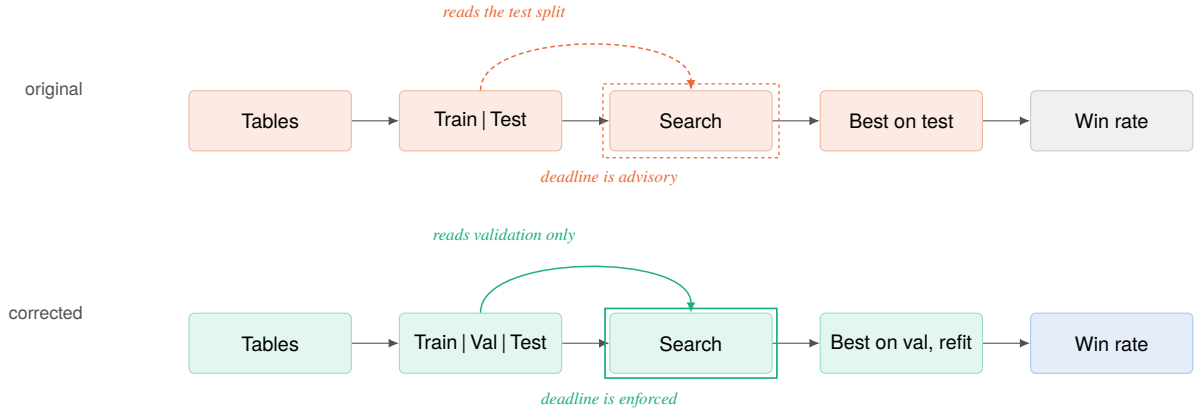

\centering
\archfigure
\caption{The protocol, before and after. Under the original protocol the search
reads the split it will be scored on, and its deadline is a check between
candidates rather than a limit, so the reported number is the best of many
test-set evaluations produced by a system that outran its budget. Under the
corrected protocol selection reads a validation split carved from the training
data, the selected model is refitted and the test split is scored exactly once,
and the deadline is enforced from outside the search. \S\ref{sec:corrected}
measures what each of those two changes is worth.}
\label{fig:architecture}
\end{figure}

Figure~\ref{fig:architecture} shows the two protocols side by side.

None of the individual observations here is novel. Selection bias in model
comparison is fifteen years old in this exact form \citep{cawley2010overfitting}
and much older in statistics; adaptive reuse of a holdout set is the subject of a
well-developed literature \citep{dwork2015reusable,blum2015ladder}; the
importance of controlling compute when comparing search methods is a standing
complaint. What we think is worth reporting is the combination: how large the
distortion is in the specific regime where informal AutoML comparisons are made,
how completely it is hidden by a conventional results table, and how easily a
system can be built that wins a benchmark for reasons that have nothing to do
with its search being good.

\section{Related Work}
\label{sec:related}

\paragraph{AutoML systems for tabular data.}
The systems we compare against sit at two ends of a design axis. \flaml{}
\citep{wang2021flaml} is built for exactly the regime studied here: it searches
over a small set of gradient-boosting and linear learners with a cost-aware
strategy that prefers cheap configurations early, which makes it strong when the
budget is measured in seconds. \ag{} \citep{erickson2020autogluon} takes the
opposite position --- fit a fixed portfolio of strong models, then stack and
ensemble them --- and is generally the stronger system when given hours. Earlier
work established the search-based framing: \textsc{Auto-WEKA}
\citep{thornton2013autoweka} and \textsc{auto-sklearn}
\citep{feurer2015autosklearn,feurer2022autosklearn2} cast pipeline construction
as combined algorithm selection and hyperparameter optimization, with
\textsc{auto-sklearn} adding meta-learned warm starts and ensemble selection
\citep{caruana2004ensemble}; \textsc{TPOT} \citep{olson2016tpot} searches
pipelines genetically; H2O AutoML \citep{ledell2020h2o} emphasises stacked
ensembles. Surveys \citep{hutter2019automl,zoller2021benchmark} and the
hyperparameter-optimization literature \citep{bischl2023hpo} cover the space in
depth. That the winning models on tabular data are almost always gradient-boosted
trees rather than neural networks is by now well established
\citep{shwartzziv2022tabular,grinsztajn2022why}, and is consistent with what all
three systems select here (\S\ref{sec:naive}).

\paragraph{Benchmarking AutoML.}
The reference protocol is the OpenML AutoML Benchmark
\citep{gijsbers2019amlb,gijsbers2024amlb}, which fixes datasets, folds, budgets
and a containerised execution environment, and reports results at 1-hour and
4-hour budgets on 10-fold cross-validation. Its underlying infrastructure ---
OpenML \citep{vanschoren2014openml} and its curated benchmarking suites
\citep{bischl2021suites} --- is what we draw datasets from, though we do not use
its harness. AMLB's design choices are, in retrospect, a list of the things we
got wrong: it never lets a system see the evaluation fold during search, it
enforces budgets with an external timeout, and it treats framework wall-clock as
a reported quantity rather than an assumption. Our contribution is not to
propose an alternative but to measure what the deviations cost.

\paragraph{Selection bias and holdout reuse.}
\citet{cawley2010overfitting} is the direct antecedent of the effect we measure:
when model selection and performance estimation use the same data, the reported
performance is biased upward, and the bias grows with the number of
configurations compared. The adaptive-data-analysis literature formalises the
same phenomenon when a holdout set is queried repeatedly
\citep{dwork2015preserving,dwork2015reusable}, and \citet{blum2015ladder} give a
leaderboard mechanism that bounds the damage. Empirically, \citet{recht2019imagenet}
and \citet{recht2018cifar} construct fresh test sets to measure how much
community-scale test-set reuse has inflated reported accuracy, and
\citet{roelofs2019metaanalysis} conduct the corresponding meta-analysis over
Kaggle competitions. Our setting is the single-run, single-machine version of
this: not a community adaptively overfitting a public benchmark over years, but
one search loop adaptively overfitting one test split over 60 seconds. The
mechanism is identical and the arithmetic is more transparent, since the number
of selection events is something we log.

\paragraph{Evaluating systems under a resource budget.}
The failure mode in \S\ref{sec:confound-time} --- a budget that is passed to
several systems and honoured by only some of them --- is familiar from systems
research, where a service-level objective is meaningless unless something
enforces it. Our own prior work on inference and serving systems repeatedly ran
into the same requirement from the other side: adaptive model partitioning under
an edge latency budget \citep{zhang2025amp4ec}, carbon- and energy-aware
inference where the budget is measured in emissions rather than seconds
\citep{zhang2026carbonedge}, GPU allocation for concurrent agents in serverless
environments \citep{zhang2025gpualloc}, admission control that refuses work when
an agent's memory budget is exhausted \citep{zhang2026memory}, and
production-scale serving where the budget is a contractual quantity
\citep{zhang2025bdaas}. Predictive autoscaling with explicit uncertainty
quantification \citep{zhang2025aapa} and simulator-based evaluation of
autoscaling policies \citep{zhang2025kiss} both exist because measuring a policy
under a budget requires controlling the environment that the budget is denominated
in --- a discipline the AutoML comparison in this paper did not apply to itself.
The design choice that matters at the model level is analogous: architectural
shortcuts are judged against a fixed evaluation harness, not against whichever
configuration happened to score best \citep{zhang2025readout}.

\paragraph{Optimising against a proxy.}
Selecting on the test set is one instance of a general pattern: a search process
optimises whatever signal it is scored against, including the noise in it. The
closest study to ours in spirit asks when a learned controller genuinely beats a
well-calibrated simple baseline on adaptive resource control, and finds the
answer depends heavily on how the comparison is set up
\citep{zhang2026calibrated} --- the same question this paper asks of AutoML search
against a fixed model pool. Reward overoptimisation in RLHF is the same failure
with a learned proxy in place of a held-out split, and can be detected by
watching for divergence between the proxy and outcomes in the world
\citep{zhang2026evalstop}. Related evaluation-integrity concerns appear when a
deployed agent memorises its evaluation inputs \citep{zhang2026memorization},
when explanation quality is measured only under favourable conditions
\citep{zhang2026robustexplain,zhang2026trust}, and when an orchestration policy
is scored against constraints it also gets to choose how to satisfy
\citep{zhang2026orchestrators}. What these share with the present paper is that
the defect is invisible in the reported metric and visible only in the protocol.

\paragraph{Compute-controlled comparison.}
\citet{bergstra2012random} is the standard demonstration that a search method's
apparent advantage depends on how much compute each arm receives, and that
random search is a stronger baseline than it looks when the budget is
equalised. The corresponding hazard in our setting is that ``budget'' names a
parameter passed to three different systems that interpret it differently: two
of them treat it as a hard wall-clock limit, and ours treated it as a loop
guard. Reporting realised wall-clock alongside nominal budget --- which AMLB
does and informal comparisons rarely do --- is enough to expose this.

\paragraph{Statistical comparison over many datasets.}
We follow the standard practice of reporting per-dataset win/loss/tie counts and
mean ranks over datasets \citep{demsar2006statistical}, and use a two-sided sign
test on the paired win/loss counts rather than a mean-rank post-hoc procedure,
following the criticism of the latter by \citet{benavoli2016mean}. We report
these statistics for both the flawed and corrected protocols, which is the point:
the significance levels under the flawed protocol are extreme, and extremeness of
a $p$-value says nothing about whether the estimand is the one you wanted.

\paragraph{Calibration and prediction markets.}
The case study in \S\ref{sec:calibration} uses the Brier score
\citep{brier1950verification} and the standard reliability-diagram construction
\citep{niculescu2005predicting,guo2017calibration}. The bias it recovers ---
low-probability contracts trading above their realised frequency --- is the
favourite--longshot bias documented across parimutuel and prediction markets
\citep{thaler1988anomalies,snowberg2010favorite,wolfers2004prediction}.

\section{The system under study}
\label{sec:system}

\orc{} is unremarkable. We describe it in enough detail to make the
experiments reproducible and to make clear that nothing in its design explains
the margins reported in \S\ref{sec:naive}.

\subsection{Pipeline}

Given a table and a target column, the system runs four stages.

\begin{enumerate}
\item \textbf{Analyse.} Load the table, drop identifier-like columns, median-fill
numeric missing values, integer-code categoricals, and split 80/20. The task type
is inferred from the target: object, boolean, or at most 20 distinct values gives
classification, otherwise regression.

\item \textbf{Metric.} Accuracy for classification. For regression, mean squared
error, except that a strictly positive target with sample skewness above $1$
switches the objective to RMSLE. (In the benchmarks below we override this and
force MSE for every regression task, so that all three systems optimise and are
scored on the same quantity.)

\item \textbf{Baselines.} Fit a fixed pool --- twelve regressors and seven
classifiers in the current code, eight and seven in the version that produced
every result reported here, listed in Appendix~\ref{app:pool} --- and keep the
best. This pool is
the floor the search has to beat, and on many datasets it is also the final
answer.

\item \textbf{Search.} Spend the remaining budget proposing and evaluating
pipeline variants (Algorithm~\ref{alg:loop}), then attempt one weighted ensemble
of the best few models found.
\end{enumerate}

\subsection{Search}

The proposer is guided random search, not Bayesian optimisation and not an
evolutionary algorithm. For the first five iterations it takes one proposal from each
not-yet-tried tree family in a fixed order --- random forest, gradient boosting,
histogram gradient boosting, extra trees, XGBoost, LightGBM. With six families
installed and five forced iterations, LightGBM never gets a guaranteed slot; it
is reachable only by later random draws. After that it refines the currently winning family with
probability $\min(0.5 + 0.01t,\ 0.75)$ at iteration $t$ and otherwise samples a
family from a fixed 70/20/10 split over tree, linear, and other (KNN, AdaBoost,
SVM) models. Hyperparameters are drawn uniformly from hand-written discrete
grids. Scaling and quantile/power transforms are proposed only for the model
families that need them; tree models are always fitted on raw features. After
iteration 10, each proposal is with probability $0.15$ a fixed voting ensemble
instead: the tree families for regression, random forest plus histogram gradient
boosting for classification.

Proposals are hashed by their description string and deduplicated, so a repeated
configuration costs nothing. In the library, batches of $2p$ proposals are
generated and evaluated across $p$ threads; in the benchmark harness used below
the loop is serial (\S\ref{sec:setup}).

When the budget expires, the current code takes the best three models of
distinct classes found during the search, weights them by inverse error (for
minimised metrics) or by score (for maximised metrics), fits a soft-voting
ensemble, and keeps it if it beats the best single model. This step was added
after the runs reported here and is absent from every result outside
\S\ref{sec:splice}, where it is the change the re-run failed to measure.

\begin{algorithm}[t]
\caption{\orc{} search loop (benchmark configuration)}
\label{alg:loop}
\begin{algorithmic}[1]
\Require training data $D_{\text{tr}}$, evaluation data $D_{\text{ev}}$, budget $B$, metric $m$
\State $t_0 \gets \textsc{now}()$;\quad $s^\star \gets \varnothing$
\For{each baseline $f$ in the fixed pool}
  \State \textbf{if} $\textsc{now}() - t_0 \ge B$ \textbf{then break}
  \State $s \gets m(f(D_{\text{tr}}), D_{\text{ev}})$;\quad
         \textbf{if} $s$ improves $s^\star$ \textbf{then} $s^\star \gets s$
\EndFor
\While{$\textsc{now}() - t_0 < B$} \Comment{checked here only --- see \S\ref{sec:confound-time}}
  \State $p \gets \textsc{Propose}(s^\star, \text{best family}, t)$
  \State $s \gets m(p(D_{\text{tr}}), D_{\text{ev}})$;\quad
         \textbf{if} $s$ improves $s^\star$ \textbf{then} $s^\star \gets s$
\EndWhile
\State $s^\star \gets \max(s^\star, \textsc{WeightedEnsemble}(\text{top-3 distinct classes}))$
\State \Return $s^\star$
\end{algorithmic}
\end{algorithm}

\subsection{The defect, stated plainly}

In the code as published and as benchmarked, $D_{\text{ev}}$ in
Algorithm~\ref{alg:loop} \emph{is the test split}. Line 4 and line 8 fit on
training data and score on test data; line 10 returns the maximum of those test
scores. There is no validation split anywhere in the loop --- the repository
contains a \texttt{splitter} module that produces one, but the search never calls
it. The number the system prints, and the number recorded in every result file in
\S\ref{sec:naive}, is therefore
\begin{equation}
\label{eq:max}
\widehat{s}_{\text{reported}}
= \max_{k=1,\dots,K} \; \widehat{s}_k^{\,\text{test}},
\end{equation}
where $K$ is the number of candidates the budget allowed and each
$\widehat{s}_k^{\,\text{test}}$ is an unbiased but noisy estimate of candidate
$k$'s generalisation performance computed on the same held-out sample.
\S\ref{sec:confound-selection} works out what that maximum is worth.

\subsection{What \orc{} is not}

It performs no neural architecture search, no stacking, no meta-learning across
datasets, no cross-validation, no feature engineering beyond imputation and
categorical coding, and no early stopping of individual fits. It has no learned
priors, despite what the project's README claims about learning ``where to look
next'' from prior benchmark runs --- no such mechanism is implemented. The
implementation is \texttt{1{,}661} lines of Python across \texttt{19} files, of
which the search loop and proposer account for \texttt{754}. Nothing here is
clever. That is the point.

\section{Experimental setup}
\label{sec:setup}

\subsection{Datasets}

We use \NData{} OpenML datasets \citep{vanschoren2014openml}: \NCls{}
classification tasks and \NReg{} regression tasks. They were drawn from OpenML
study suites 218 (the AutoML benchmark suite), 99 (CC18) and 269 (regression)
\citep{bischl2021suites}, filtered to between 50 and 100{,}000 instances, at most 500
features, under 30\% missing values, at most 30 classes, and active status, then
sorted by instance count. The \NData{} are those on which every framework
produced a score without error in the 30-second sweep, in which \TwoErr{} of
\TwoDedup{} attempted datasets failed to load, had no usable target, or had
fewer than 50 usable rows. The 60-second sweep then ran on those \NData{} and
none failed.

The resulting collection is larger than the AMLB suites and correspondingly less
curated. Sample counts run from \SampMin{} to \SampMax{} (median \SampMed{});
feature counts from \FeatMin{} to \FeatMax{} (median \FeatMed{}). \DupNames{}
dataset names appear more than once, because OpenML assigns distinct IDs to
different versions of the same underlying data; we did not deduplicate these, so
the \NData{} tasks are not fully independent. All per-dataset results are
released, so any reader who prefers a stricter selection can recompute the
aggregates.

\subsection{Protocol}

Every framework sees exactly the same data. For each dataset we take the OpenML
default target, integer-code categorical columns and the target if it is
categorical, median-impute numeric columns, drop all-missing columns and
rows with a missing target, cast to \texttt{float64}, and split once, 80/20,
with \texttt{random\_state=42}. There is no cross-validation and no repetition:
each dataset contributes one comparison. Classification is scored by accuracy
and regression by mean squared error, with the framework's own internal
objective set to match.

A dataset is won by the framework with the best score, with ties declared when
scores fall within $10^{-6}$; ties are reported as their own category rather than
split. We report win counts, pairwise head-to-head counts, mean rank over
datasets \citep{demsar2006statistical}, and a two-sided sign test on paired
wins and losses \citep{benavoli2016mean}.

\subsection{Frameworks and budgets}

\flaml{} 2.5.0 is called with its default estimator list, \texttt{seed=42}, and
\texttt{metric} set to accuracy or MSE. \ag{} 1.5.0 is called with
\texttt{presets=\allowbreak"medium\_quality"} and \texttt{num\_gpus=0}. One caveat: the
environment lacked the optional \texttt{fastai} dependency, so \ag{} skipped its
\texttt{NeuralNetFastAI} model on every dataset and fitted its remaining
portfolio; this can only have hurt \ag{}, and we flag it wherever \ag{} numbers
are quoted. \orc{} is run through the harness in \texttt{experiments/}, which reimplements
the loop serially --- no proposal batching, no dedup cache --- but keeps the same
proposer and baseline pool. Note that the post-search weighted ensemble and four
of the twelve regression baselines described in \S\ref{sec:system} were added
after these runs; every result outside \S\ref{sec:splice} therefore comes from an
eight-model regression pool with no ensembling step. One caveat on the code state: the classification and
30-second results predate commit \texttt{c9e337a}, which added four regression
baselines and the post-search ensemble; only the regression re-run discussed in
\S\ref{sec:splice} used the later code. Since that re-run is excluded from
Table~\ref{tab:main}, every number in it comes from the earlier state.

Two budgets are studied: a 30-second two-way comparison against \flaml{}, run
serially, and a 60-second three-way comparison, run with four dataset-level
worker processes. The distinction matters for \S\ref{sec:confound-time}: with
four workers each fitting models that request every core, the machine is
oversubscribed, and a framework that enforces its budget internally responds by
doing fewer trials while a framework that does not responds by taking longer.

\subsection{Environment}

All runs are CPU-only on an NVIDIA DGX Spark: 20 ARM64 cores (Cortex-X925 /
Cortex-A725, \texttt{aarch64}), 121\,GB of unified memory, Linux 6.14. Python
3.12.3, scikit-learn 1.7.2 \citep{pedregosa2011scikit}, NumPy 2.3.5, pandas
2.3.3, XGBoost 3.2.0 \citep{chen2016xgboost}, LightGBM 4.6.0
\citep{ke2017lightgbm}, CatBoost 1.2.10 \citep{prokhorenkova2018catboost},
\flaml{} 2.5.0, \ag{} 1.5.0.

\subsection{What is released}

The repository accompanying this paper contains the per-dataset JSONL for every
run reported here, the harnesses that produced them, and two scripts that
regenerate every number and every figure in this paper directly from those
files: \texttt{paper/figures/make\_numbers.py} emits the macro file that the
text is written against, so no quantity in the prose is transcribed by hand, and
\texttt{paper/figures/make\_figures.py} emits the figures. The
protocol-corrected harness of \S\ref{sec:corrected} is released as
\texttt{experiments/fair\_benchmark.py}.

\section{The result as originally obtained}
\label{sec:naive}

This section reports what the original protocol produced. Nothing here is
corrected; \S\ref{sec:confounds} and \S\ref{sec:corrected} do that. We present it
in full because the argument of the paper is that a table of this kind can look
entirely healthy while resting on an invalid estimand.

\subsection{Three-way comparison at 60 seconds}

\begin{figure}[t]
\centering
\includegraphics[width=\linewidth]{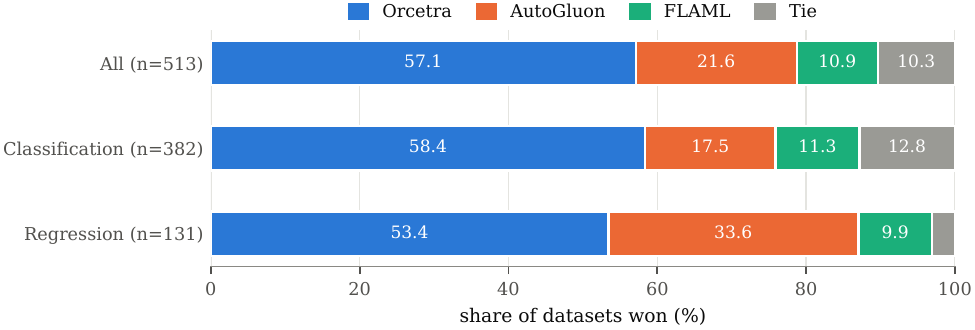}
\caption{Win-rate composition under the original protocol at a 60-second nominal
budget. Each bar is a partition of the datasets in that stratum.}
\label{fig:winrate}
\end{figure}

Table~\ref{tab:main} and Figure~\ref{fig:winrate} give the headline.
\orc{} wins \WinOrcAll{}\% of the \NData{} datasets, against \WinAgAll{}\% for
\ag{} and \WinFlAll{}\% for \flaml{}, with \WinTieAll{}\% ties. The margin is
present in both task types, wider on classification (\WinOrcCls{}\% against
\WinAgCls{}\%) than on regression (\WinOrcReg{}\% against \WinAgReg{}\%).

These \NData{} datasets come from one sweep, run under one machine load. That
qualification turns out to matter (\S\ref{sec:splice}).

\begin{table}[t]
\centering
\begin{threeparttable}
\caption{Win rates under the original protocol, 60-second nominal budget,
\NData{} OpenML datasets. Percentages of datasets on which each framework
achieved the strictly best score; ties are counted separately.
\textbf{These numbers are not valid} --- see \S\ref{sec:confounds}.}
\label{tab:main}
\small
\begin{tabular}{@{}lrrrrr@{}}
\toprule
& $n$ & \orc{} & \ag{}\tnote{a} & \flaml{} & Tie \\
\midrule
All datasets   & \NData{} & \textbf{\WinOrcAll{}\%} & \WinAgAll{}\% & \WinFlAll{}\% & \WinTieAll{}\% \\
Classification & \NCls{}  & \textbf{\WinOrcCls{}\%} & \WinAgCls{}\% & \WinFlCls{}\% & \WinTieCls{}\% \\
Regression     & \NReg{}  & \textbf{\WinOrcReg{}\%} & \WinAgReg{}\% & \WinFlReg{}\% & \WinTieReg{}\% \\
\bottomrule
\end{tabular}
\begin{tablenotes}\footnotesize
\item[a] \ag{} ran without its optional \texttt{fastai} neural model
(\S\ref{sec:setup}); its numbers are a lower bound.
\end{tablenotes}
\end{threeparttable}
\end{table}

The pairwise picture (Table~\ref{tab:h2h}) is stronger still, because it removes
the three-way ties. Head to head, \orc{} beats \flaml{} on \HOrcFlWp{}\% of
datasets and loses on \HOrcFlLp{}\%, and beats \ag{} on \HOrcAgWp{}\% against
\HOrcAgLp{}\%. Two-sided sign tests on the paired counts give
$p = \HOrcFlP{}$ and $p = \HOrcAgP{}$. \ag{} beats \flaml{} on \HAgFlWp{}\% of
datasets, which is the expected ordering between those two systems and a useful
sanity check that the harness is not simply broken.

\begin{table}[t]
\centering
\caption{Pairwise comparisons and aggregate quality under the original protocol,
\NData{} datasets, 60-second nominal budget. Mean rank is over all three systems
(1 = best). Relative MSE is each system's MSE divided by the best MSE on that
dataset, aggregated over the \NRegRel{} regression tasks whose best MSE is
strictly positive.}
\label{tab:h2h}
\small
\begin{tabular}{@{}lrrr@{}}
\toprule
& \orc{} & \ag{} & \flaml{} \\
\midrule
\multicolumn{4}{@{}l}{\emph{Head-to-head (win / loss / tie)}}\\
\quad vs.\ \flaml{}   & \HOrcFlW{} / \HOrcFlL{} / \HOrcFlT{} & \HAgFlW{} / \HAgFlL{} / \HAgFlT{} & --- \\
\quad vs.\ \ag{}      & \HOrcAgW{} / \HOrcAgL{} / \HOrcAgT{} & --- & \HAgFlL{} / \HAgFlW{} / \HAgFlT{} \\[3pt]
\multicolumn{4}{@{}l}{\emph{Aggregate}}\\
\quad Mean rank                       & \textbf{\RankOrc{}} & \RankAg{} & \RankFl{} \\
\quad Mean accuracy (classification)  & \textbf{\MeanAccOrc{}} & \MeanAccAg{} & \MeanAccFl{} \\
\quad Median relative MSE (regression)& \textbf{\RelMseOrc{}} & \RelMseAg{} & \RelMseFl{} \\
\quad Geometric-mean relative MSE     & \textbf{\RelMseGeoOrc{}} & \RelMseGeoAg{} & \RelMseGeoFl{} \\
\bottomrule
\end{tabular}
\end{table}

\subsection{Are the datasets independent?}

OpenML assigns separate IDs to different versions of the same underlying data, and
\DupNames{} names recur in our collection, so the \NData{} rows are not
\NData{} independent units. Keeping only one record per name --- the largest by
sample count --- leaves \UniqN{} datasets (\UniqNCls{} classification,
\UniqNReg{} regression). \orc{}'s share barely moves, from \WinOrcAll{}\% to
\UniqOrc{}\%; the other two shift by a couple of points in opposite directions,
\ag{} to \UniqAg{}\% and \flaml{} to \UniqFl{}\%, with \UniqTie{}\% ties. Head to head on the deduplicated collection, \orc{} beats
\flaml{} \UniqHOrcFlW{}--\UniqHOrcFlL{} ($p = \UniqHOrcFlP{}$) and \ag{}
\UniqHOrcAgW{}--\UniqHOrcAgL{} ($p = \UniqHOrcAgP{}$). Duplication is not what
produces the result, and we use the full collection below.

\subsection{Two-way comparison at 30 seconds}

\begin{figure}[t]
\centering
\includegraphics[width=0.48\linewidth]{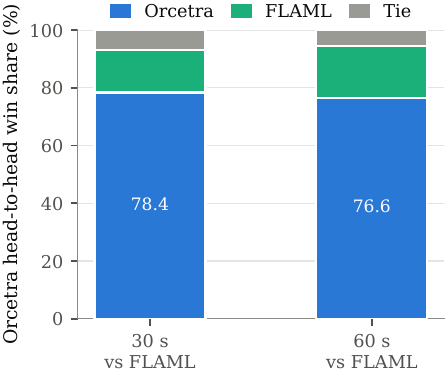}
\caption{\orc{} versus \flaml{} head to head, at both budgets, under the original
protocol. Agreement across budgets was originally read as confirmation; it is equally
consistent with a bias present at both budgets.}
\label{fig:budget}
\end{figure}

The earlier and simpler experiment compares \orc{} against \flaml{} alone at a
30-second budget, run serially rather than with parallel workers. \orc{} wins
\TwoOrc{}\% of \TwoN{} datasets, \flaml{} wins \TwoFl{}\%, and \TwoTie{}\% tie.
Figure~\ref{fig:budget} places this beside the corresponding head-to-head
restriction of the 60-second run; the two agree closely, which at the time we
read as evidence that the result held across budgets.

\subsection{Which models won}

The selected models are unremarkable and consistent with the tabular literature
\citep{grinsztajn2022why}. \orc{} most often ends on
\texttt{\TopModelOrc{}} (\TopModelOrcN{} of \NData{} datasets), followed by
random forests and logistic regression. \flaml{} selects \texttt{\TopModelFl{}}
on \TopModelFlPct{}\% of datasets. \ag{} returns a weighted ensemble on
\AgEnsemblePct{}\% --- which is what it is designed to do, and is the first hint
that the comparison is not measuring what it appears to: \ag{} is being scored on
an honestly stacked ensemble while \orc{} is being scored on the luckiest of many
single models.

\subsection{An internal check that also looked good}

An earlier single-framework sweep, over \SweepN{} OpenML datasets
(\SweepNCls{} classification, \SweepNReg{} regression) at a nominal 30-second
search budget, compared \orc{}'s search against its own baseline pool rather
than against another framework. The search improved on the best baseline on
\SweepWinPct{}\% of datasets, with a median improvement of \SweepImpMed{}\% and a
median of \SweepIterMed{} search iterations per dataset. This is the number that
made the result feel solid: the search was apparently finding real gains over a
strong fixed pool, on most datasets, quickly.

It suffers from both defects at once. The baseline score and the search score are
each maxima over test-set evaluations, and the search takes its maximum over far
more of them --- a median of \SweepIterMed{} candidates against a pool of at most
twelve --- so the comparison is rigged in the search's favour by construction.
The same sweep also shows the budget defect in its rawest form: against a
30-second budget the median run took \SweepTimeMed{}\,s and the longest took
\SweepTimeMax{}\,s.

\section{What actually produced the margin}
\label{sec:confounds}

\subsection{Selection on the test set}
\label{sec:confound-selection}

Equation~\ref{eq:max} says the reported score is a maximum over $K$ test-set
evaluations. What does that maximum buy?

Suppose the $K$ candidates a 60-second budget reaches have comparable true
accuracy $\mu$, and that the test-set estimate of candidate $k$ is
$\widehat{s}_k = \mu + \varepsilon_k$ with
$\varepsilon_k \sim \mathcal{N}(0, \tau^2)$ for some scale $\tau$. Then the
expected maximum exceeds $\mu$ by approximately $\tau\sqrt{2\ln K}$.

Two things about this deserve care, because both cut against us.

The first is the choice of $\tau$. The obvious substitution --- the marginal
standard error of one model's accuracy, $\sigma = \sqrt{p(1-p)/m}$ on a test set
of $m$ rows --- is \emph{not} the right scale, and it is too large. All $K$
candidates are scored on the same held-out rows, so $\widehat{s}_k$ decomposes
into a term reflecting which rows happened to land in the test split, common to
every candidate, and a term reflecting where this particular model disagrees with
the others. The common term shifts all $K$ estimates together and cancels out of a maximum;
only the second term is available for the maximum to exploit. How large that
second term is depends on how much the candidates disagree, which is an empirical
question and not one $\sigma$ answers. Rather than estimate it, we measure the
realised bias directly in \S\ref{sec:curve}, where the corrected run logs every
candidate's test score. Here we substitute $\sigma$ and read the result strictly
as an upper bound.

The second is $K$, and it turns out not to matter much. The factor
$\sqrt{2\ln K}$ moves only from \SqrtTwoLnKLo{} to \SqrtTwoLnKHi{} as $K$ ranges
over \KSelLo{}--\KSelHi{}, so an error of a factor of three in the candidate
count changes the estimate by under ten percent. Nothing below turns on the exact
value.

With those caveats, the arithmetic: on our classification tasks the median test
split is \TestSizeMed{} rows and the mean accuracy is \MeanAccBar{}, giving
$\sigma \approx \SigmaMedPts{}$ accuracy points; the search evaluates on the
order of $K = \KSel{}$ candidates in 60 seconds, and
$\sqrt{2\ln \KSel{}} = \SqrtTwoLnK{}$. Computed per dataset rather than at the
median, the predicted inflation is \InflPredMed{} accuracy points (interquartile
range \InflPredQone{}--\InflPredQthree{}) --- again, an upper bound.

Now compare that against the margins the experiment is resolving. Across the
\NCls{} classification datasets, \orc{}'s median lead over the better of \flaml{}
and \ag{} is \MarginMed{} accuracy points. On \MarginCloseP{}\% of datasets the
absolute gap between \orc{} and the best competitor is smaller than the inflation
predicted for that dataset. Among the datasets \orc{} wins, the median winning
margin is \WinMarginMed{} points, and \WinMarginCloseP{}\% of those wins are by
less than the median predicted inflation.

The margins being decided sit an order of magnitude below the upper bound on the
bias. That does not by itself establish that the result is an artifact --- the
bound may be loose by exactly that order of magnitude, which is why
\S\ref{sec:corrected} measures the effect instead of inferring it. What the
calculation does establish, before any re-run, is that the experiment was never
powered to resolve the margins it was reporting. This is what makes the defect
dangerous: it does not produce implausible numbers. It produces a plausible
one-to-two-point edge, sustained across hundreds of datasets, with the tiny
per-dataset margins that a real but modest improvement would also produce.

Two further consequences are worth stating. First, the bias grows with the
budget, since $K$ grows with the budget: a system with this defect will appear to
scale better with compute than it does. Second, it interacts with the tie rule.
Ties are declared at $10^{-6}$, far below the noise floor, so near-ties are
resolved as wins --- and a maximum-over-$K$ estimator wins near-ties
systematically.

\subsection{Budgets that are checked but not enforced}
\label{sec:confound-time}

\begin{figure}[t]
\centering
\includegraphics[width=\linewidth]{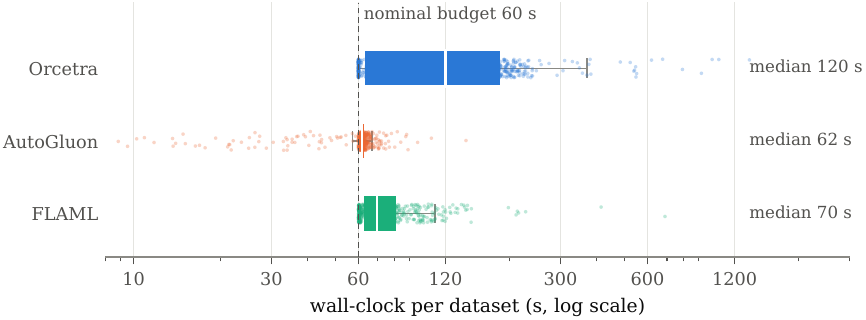}
\caption{Wall-clock actually consumed per dataset at a nominal 60-second budget.
\ag{} holds the limit closely; \flaml{} overshoots modestly (median
\TimeFlMed{}\,s, maximum \TimeFlMax{}\,s) because its budget governs search
rather than the final fit; \orc{} checks the deadline only between candidates and
runs to \TimeOrcMax{}\,s. Points are individual datasets; boxes are quartiles
with whiskers at 1.5\,IQR.}
\label{fig:walltime}
\end{figure}

Figure~\ref{fig:walltime} shows what a deadline that never preempts costs. At a
nominal 60-second budget,
\orc{} consumed a median of \TimeOrcMed{}\,s per dataset (mean \TimeOrcMean{}\,s,
90th percentile \TimeOrcPninety{}\,s, maximum \TimeOrcMax{}\,s), against
\TimeFlMed{}\,s for \flaml{} and \TimeAgMed{}\,s for \ag{}. It exceeded its
budget on \OverBudgetPct{}\% of datasets and more than doubled it on
\OverBudgetTwoX{}\%. Neither competitor is perfectly obedient either --- \flaml{}
runs past 60\,s on every dataset and to \TimeFlMax{}\,s at worst, since its
budget governs the search rather than the final fit --- but the overshoot is
bounded, where \orc{}'s is not. Per dataset, the median ratio of \orc{}'s wall-clock to
\flaml{}'s is \RatioOrcFlMed{}$\times$ and to \ag{}'s \RatioOrcAgMed{}$\times$.

The distortion is amplified by the execution setup. The 60-second run used four
dataset-level worker processes on a 20-core machine, with every fit requesting
all cores. Under that contention, \flaml{} and \ag{} keep their wall-clock and
give up trials; \orc{} keeps its trials and gives up wall-clock. The nominal
budget is the same string passed to all three, and it means three different
things.

\paragraph{Isolating the effect.} Restricting to datasets on which \orc{}
consumed no more wall-clock than a competitor gives a compute-matched slice
(Table~\ref{tab:matched}). Matching against \flaml{} leaves \MatchFlN{} datasets
and drops \orc{} from \WinOrcAll{}\% to \MatchFlOrc{}\%; matching against \ag{}
leaves only \MatchAgN{} datasets --- \orc{} rarely finished as fast as \ag{} ---
and drops it to \MatchAgOrc{}\%; requiring both leaves \MatchBothN{} datasets and
\MatchBothOrc{}\%. These slices are not unbiased estimates of a properly
budget-controlled comparison: conditioning on \orc{} having been fast selects
for easy datasets, and the smaller slices are noisy. They establish direction and
rough size, not a corrected number.

\begin{table}[t]
\centering
\caption{Compute-matched slices of the 60-second run. Conditioning on \orc{}
having used no more wall-clock than the named competitor. These are diagnostic,
not a corrected result: the conditioning is not independent of dataset
difficulty.}
\label{tab:matched}
\small
\begin{tabular}{@{}lrrrrr@{}}
\toprule
Restriction & $n$ & \orc{} & \ag{} & \flaml{} & Tie \\
\midrule
None (Table~\ref{tab:main})                    & \NData{}     & \WinOrcAll{}\%  & \WinAgAll{}\%  & \WinFlAll{}\%  & \WinTieAll{}\% \\
$t_{\orc{}} \le t_{\flaml{}}$                  & \MatchFlN{}  & \MatchFlOrc{}\% & \MatchFlAg{}\% & \MatchFlFl{}\% & \MatchFlTie{}\% \\
$t_{\orc{}} \le t_{\ag{}}$                     & \MatchAgN{}  & \MatchAgOrc{}\% & \MatchAgAg{}\% & \MatchAgFl{}\% & \MatchAgTie{}\% \\
$t_{\orc{}} \le \min(t_{\flaml{}}, t_{\ag{}})$ & \MatchBothN{}& \MatchBothOrc{}\% & \MatchBothAg{}\% & \MatchBothFl{}\% & \MatchBothTie{}\% \\
\bottomrule
\end{tabular}
\end{table}

\subsection{A third defect, found while writing this paper}
\label{sec:splice}

The two defects above were known before we started writing. Auditing the result
files for this paper turned up a third, and it is the one we would most easily
have published without noticing.

The result files in the repository include a later regression-only re-run,
stored beside the original sweep with nothing to mark which supersedes which.
Loading every \texttt{multi\_framework\_*.jsonl} and deduplicating by dataset ID
--- the obvious thing to do, and what our own first pass did --- silently splices
the two, turning Table~\ref{tab:main}'s
\WinOrcAll{}\,/\,\WinAgAll{}\,/\,\WinFlAll{}\,/\,\WinTieAll{}\% into
\MergedOrc{}\,/\,\MergedAg{}\,/\,\MergedFl{}\,/\,\MergedTie{}\%.

The re-run was made to measure a code change --- a post-search weighted ensemble
had been added --- and on its face it worked: \orc{}'s regression wins went from
\SpliceOldOrc{} to \SpliceNewOrc{} of \SpliceN{} datasets, and \ag{}'s fell from
\SpliceOldAg{} to \SpliceNewAg{}.

The re-run measured nothing of the kind.

\begin{itemize}
\item The ensemble never won. It is the selected model on \SpliceEnsembleN{} of
the \SpliceN{} regression datasets.
\item \orc{}'s own score is bit-identical on \SpliceSame{} of \SpliceN{}
datasets, better on \SpliceBetter{}, and \emph{worse} on \SpliceWorse{}.
\item What changed is the competitors. \ag{} scored worse in the re-run on
\SpliceAgWorse{} of \SpliceN{} datasets and \flaml{} on \SpliceFlWorse{}.
\item \orc{}'s median wall-clock over the same datasets rose from
\SpliceTimeOld{}\,s to \SpliceTimeNew{}\,s, which is the tell: the machine was
more heavily loaded during the re-run, so the two frameworks that respect a
wall-clock budget did fewer trials while the one that does not simply took
longer.
\end{itemize}

The apparent \SpliceOldOrc{}$\to$\SpliceNewOrc{} improvement is a contention
artifact. We therefore report Table~\ref{tab:main} from the single internally
consistent sweep and treat \MergedOrc{}\% as what the splice produces rather
than as a result. The repository's README, which predates the re-run, reports
the unspliced figure and is correct; the hazard is not that anyone published the
merged number but that nothing in the file layout stops the next person from
computing it.

The general rule is narrower than ``do not merge runs'': results merged across
executions are comparable only if every framework in them was re-measured
together. Here only one was.

\subsection{The confounds are separable}

The 30-second two-way experiment is a useful control, because it was run
serially: one dataset at a time, no contention. There \orc{}'s median wall-clock
was \TwoTimeOrcMed{}\,s against \flaml{}'s \TwoTimeFlMed{}\,s, a median
per-dataset ratio of \TwoRatioMed{}$\times$. The medians match; the tails do not,
and \orc{} still ran long on some datasets. But at the median, compute was
matched --- and \orc{} still won \TwoOrc{}\% of datasets.

So the budget defect cannot be the whole story. The 30-second experiment has no
budget defect and reports an inflated margin regardless. What the two experiments
share is selection on the test set, and that defect cannot be removed by
re-slicing the results --- only by re-running, which is what
\S\ref{sec:corrected} does.

\section{The corrected protocol}
\label{sec:corrected}

\subsection{Design}

We re-ran the three-way comparison with both defects fixed and everything else
held fixed --- same datasets, same preprocessing, same split seed, same budget,
same framework versions, same machine.

\paragraph{Selection moves off the test set.} The training split is divided again,
80/20, into a search-training set and a validation set. Every baseline and every
proposal is fitted on the search-training set and scored on validation. When the
budget expires, the single model with the best validation score is refitted on
the full training split and evaluated once on the test split. That test
evaluation is the reported number. The test set is used exactly once per dataset,
as it is for \flaml{} and \ag{}.

\paragraph{The budget becomes an external deadline.} The whole search runs in a
child process. The child stops proposing at the budget; the parent kills it outright
if it has not finished within budget plus a fixed grace period. This bounds the
overshoot rather than eliminating it --- a single candidate can still run to the
end of the grace window --- so we report realised search time rather than
asserting the budget was kept, and a killed run yields no model at all rather
than a partial one. The final refit is allowed to complete outside the budget
and its cost is reported separately.

\paragraph{The old number is recorded alongside the new one.} During the same
search, each candidate is fitted once and predicted twice --- on validation, which
drives selection, and on test, which drives nothing. For search candidates this
costs one extra prediction, not an extra fit. When a \emph{baseline} wins on
validation one extra fit is needed, since the baseline pool is scored through a
different code path; that happens on a substantial minority of datasets and is
charged to the search clock, making the reported search times conservative. The
run yields three estimands:

\begin{description}[leftmargin=2.4em,style=nextline,itemsep=2pt,topsep=3pt]
\item[$A$ --- select on test] the running maximum of the test scores, which
applies the original protocol's \emph{selection rule} to this run. It is not a
replay of the original experiment: $K$, the training-set size, the preemption and
the core allocation all differ. It also maxes over the search candidates and the
val-selected baseline rather than the whole baseline pool, so it understates the
original maximum, making $A - B$ conservative.
\item[$B$ --- select on validation, same training data] the test score of the
model the validation split chose, fitted on the same 64\% of the data as every
candidate in $A$.
\item[$C$ --- select on validation, refit] that same model refitted on the full
80\% training split and evaluated once. This is the number we report, and it is
the one comparable to \flaml{} and \ag{}, which also train on the full split.
\end{description}

\noindent
$A$ and $B$ differ \emph{only} in the selection rule, with training-set size,
candidate set, budget and test split all held fixed, so $A - B$ is a clean paired
measurement of what selecting on the test set is worth. $C$ additionally restores
the training data the validation split consumed.

We ran this on a random subsample of the \NData{} datasets, drawn with a fixed
seed and stratified to preserve the classification/regression ratio, with four
worker processes each pinned to five cores so that the budget means the same
thing for all three frameworks.

\subsection{Results}

\paragraph{The headline.} Over \FairN{} re-run datasets (\FairNCls{}
classification, \FairNReg{} regression), \orc{} wins \FairOrc{}\%, \flaml{}
\FairFl{}\% and \ag{} \FairAg{}\%, with \FairTie{}\% ties. On the same
datasets the original protocol reported \FairOrigOrc{}\,/\,\FairOrigAg{}\,/\,%
\FairOrigFl{}\,/\,\FairOrigTie{}\%. Fixing the protocol costs \orc{} roughly
twenty-five percentage points and erases the result: head to head it now beats
\flaml{} \FairHOrcFlW{}--\FairHOrcFlL{} ($p = \FairHOrcFlP{}$) and \ag{}
\FairHOrcAgW{}--\FairHOrcAgL{} ($p = \FairHOrcAgP{}$). Neither is
distinguishable from a coin flip. At a hard 60-second budget on matched compute,
the three systems are statistically indistinguishable on this collection, and
the margin in Table~\ref{tab:main} was an artifact.

\paragraph{Which defect did what.} Table~\ref{tab:corrected} separates the two,
and the answer is not the one the title leads with. Scoring one set of searches
three ways over \PairedN{} datasets: under the original select-on-test rule
($A$) \orc{} wins \FairPairedOracleOrc{}\%; changing nothing but the selection
rule ($B$) takes it to \FairMatchedOrc{}\%. The selection rule is therefore
worth \SelRulePts{} percentage points --- real, measured with everything else
held fixed, and far short of the twenty-five points that separate the original
and corrected headlines. Restoring the training data the validation split
consumed ($C$) gives \FairPairedHonestOrc{}\%.

The remainder belongs to the budget. What changed between the original run and
this one, beyond selection, is that the search now stops at an external deadline
and each framework gets an equal, pinned share of the machine. \flaml{}, the
system most starved by the original contention regime, is the main beneficiary:
it goes from \FairOrigFl{}\% to \FairFl{}\% on the same datasets. Unequal
compute, not peeking, did most of the damage here.

\begin{table}[t]
\centering
\caption{One set of searches on \PairedN{} datasets, scored three ways. Rows $A$
and $B$ differ only in the selection rule; row $C$ additionally refits the
selected model on the full training split, matching how \flaml{} and \ag{} use
the data. All three rows use the identical \flaml{} and \ag{} scores.}
\label{tab:corrected}
\small
\begin{tabular}{@{}llrrrr@{}}
\toprule
& Scoring of \orc{} & \orc{} & \ag{} & \flaml{} & Tie \\
\midrule
$A$ & select on test (original)              & \FairPairedOracleOrc{}\% & \FairPairedOracleAg{}\% & \FairPairedOracleFl{}\% & \FairPairedOracleTie{}\% \\
$B$ & select on validation, same training data & \FairMatchedOrc{}\% & \FairMatchedAg{}\% & \FairMatchedFl{}\% & \FairMatchedTie{}\% \\
$C$ & select on validation, refit \emph{(reported)} & \FairPairedHonestOrc{}\% & \FairPairedHonestAg{}\% & \FairPairedHonestFl{}\% & \FairPairedHonestTie{}\% \\
\bottomrule
\end{tabular}
\end{table}

\begin{figure}[t]
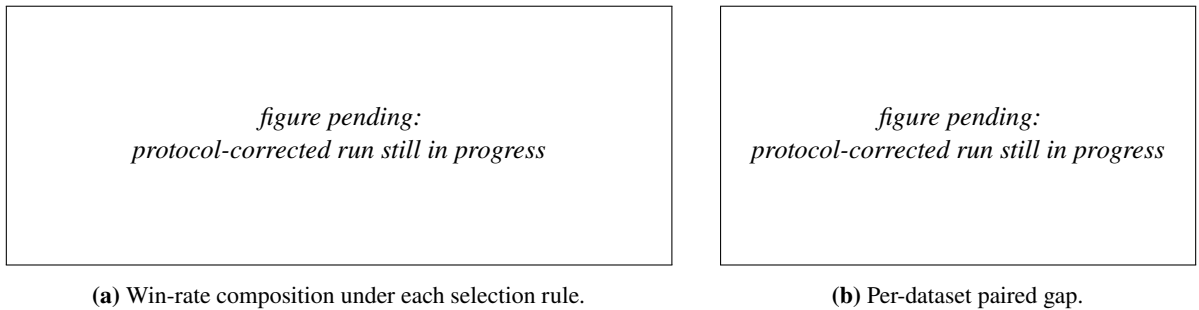

\centering
\begin{subfigure}{0.56\linewidth}
  \figorpending{fig_protocol.pdf}{\linewidth}
  \caption{Win-rate composition under each selection rule.}
  \label{fig:protocol}
\end{subfigure}\hfill
\begin{subfigure}{0.40\linewidth}
  \figorpending{fig_oracle_gap.pdf}{\linewidth}
  \caption{Per-dataset paired gap.}
  \label{fig:oraclegap}
\end{subfigure}
\caption{The cost of selecting on the test set, measured within single runs.
(\subref{fig:protocol}) one set of searches scored under all three rules;
(\subref{fig:oraclegap}) the $A$-versus-$B$ comparison per classification
dataset --- both axes come from the same candidate set fitted on the same
training data, so every point above the diagonal is the selection rule and
nothing else.}
\label{fig:corrected}
\end{figure}

The per-dataset view (Figure~\ref{fig:oraclegap}) shows why the selection effect
is smaller than the bound allowed for, and it is not that the effect is absent.
Across the \PairedNCls{} paired classification datasets the gap $A - B$ is
\emph{exactly zero} on two thirds of them: on most datasets the model the
validation split picks is also the model that maximises the test score, so
peeking buys nothing. On the remaining third it buys a great deal --- the
distribution has mean \GapClsMeanPts{} accuracy points, upper quartile
\GapClsQthreePts{}, and a maximum of \GapClsMaxPts{}. The bias is concentrated,
not diffuse, which is precisely what makes it hard to notice in an aggregate and
easy to notice in a win rate: a win rate is decided by the datasets where the
margin is thin, and those are the datasets where peeking pays.

Head to head under the reported protocol, \orc{} beats \flaml{} on
\FairHOrcFlW{} datasets and loses on \FairHOrcFlL{} ($p = \FairHOrcFlP{}$), and
beats \ag{} on \FairHOrcAgW{} and loses on \FairHOrcAgL{}
($p = \FairHOrcAgP{}$).

\subsection{The empirical selection curve}
\label{sec:curve}

\S\ref{sec:confound-selection} had to guess at the scale $\tau$ of the noise a
maximum exploits, and settled for an upper bound. The candidate traces make the
guess unnecessary. For each dataset we take the first $K'$ candidates the search
evaluated and compute two numbers: the best test score among them, which is what
select-on-test would report at that budget, and the test score of the candidate
the \emph{validation} split would have picked from the same prefix. Their
difference is the selection bias actually realised at budget $K'$, with no
distributional assumption anywhere.

\begin{figure}[t]
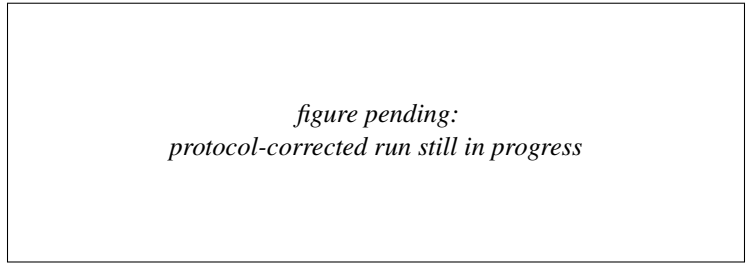

\centering
\figorpending{fig_curve.pdf}{0.62\linewidth}
\caption{Selection bias realised at budget $K'$, measured within runs on the
\TraceN{} traced classification datasets. At each $K'$ the curve is the mean
over datasets of (best test score among the first $K'$ candidates) minus (test
score of the candidate validation would have chosen from the same prefix). The
dashed line is the $\sigma\sqrt{2\ln K}$ upper bound of
\S\ref{sec:confound-selection}.}
\label{fig:curve}
\end{figure}

Averaged over the \TraceN{} classification datasets with traces, that difference
grows with budget, though not monotonically --- \TraceInflKtwo{} accuracy points
at $K' = 2$, \TraceInflKten{} at $K' = 10$, \TraceInflMax{} at
$K' = \TraceKMax{}$, with small non-monotone dips in between
(Figure~\ref{fig:curve}). Only a minority of traces are long enough to reach the
largest prefix, so the right-hand end of the curve is partly saturated shorter
traces rather than genuinely deeper searches. The qualitative claim survives:
the bias is real and it grows with the compute budget, because the budget is
what buys $K$.

The quantitative claim does not survive in the form \S\ref{sec:confound-selection}
stated it. The realised bias at the largest budget is \TraceInflMax{} accuracy
points against a $\sigma\sqrt{2\ln K}$ bound of \InflPredMed{} --- the bound
overestimates by roughly a factor of five. The two are not quite the same
estimand: the bound is the excess of a maximum over the common mean, while what
we measure is the excess of that maximum over an already well-chosen candidate,
so some of the gap is definitional. It is nonetheless the comparison a
practitioner would make, since the validation-selected model is what an honest
protocol reports. This is the concern raised in
\S\ref{sec:confound-selection} coming true: candidates are scored on the same
held-out rows, the common component cancels out of a maximum, and the marginal
standard error of a single model's accuracy is the wrong scale by about that
factor. We report it because a screen that overstates by $5\times$ is still
useful --- a margin below the bound cannot be trusted --- but it is not a
prediction, and anyone using it should read a margin \emph{above} the bound as
the only clear signal.

\subsection{What the corrected budget cost}

\orc{}'s median wall-clock falls to \FairTimeOrcMed{}\,s. The search phase alone
has median \FairSearchMed{}\,s against the 60-second budget, exceeds it on
\FairSearchOverPct{}\% of datasets, and reaches \FairSearchMax{}\,s at worst;
the parent had to kill the child on \FairKilledN{} datasets
(\FairKilledPct{}\%). Total wall-clock per dataset therefore fell from \TimeOrcMed{}\,s under the
original harness to \FairTimeOrcMed{}\,s, without the budget becoming exact ---
which is the honest description, and still leaves \orc{} the most budget-elastic
of the three systems. The search reaches a median of
\FairNEvalMed{} candidate evaluations per dataset (maximum \FairNEvalMax{}); the
$K = \KSel{}$ used in \S\ref{sec:confound-selection} is the right order but
somewhat high, which further loosens that bound.

\section{Case study: prediction-market calibration}
\label{sec:calibration}

\orc{} did not start as an AutoML tool. It started as a forecasting system for
Polymarket, a prediction market, and the model-search loop is what remained after
the domain-specific parts were removed. We include the original application for
two reasons: it is the one place where the framework is applied to something
other than a curated tabular benchmark, and the results are partly negative.

\subsection{What the data supports, and how much}

\begin{figure}[t]
\centering
\includegraphics[width=0.48\linewidth]{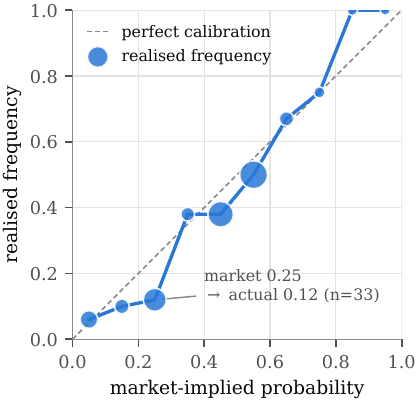}
\caption{Reliability diagram over \CalN{} resolved Polymarket contracts. Marker
area grows with the number of markets in the bin, which ranges from 2 to 50: the visually striking deviations are mostly the small bins. Points below the
diagonal are bands where the market overprices the outcome.}
\label{fig:calibration}
\end{figure}

The calibration component learns a mapping from stated probability to realised
frequency by binning resolved contracts. Fitted on \CalN{} resolved Polymarket
markets across \CalNBins{} bins, it produces the reliability diagram in
Figure~\ref{fig:calibration}, which leans in the direction of the
favourite--longshot bias documented in parimutuel and prediction markets for
decades \citep{thaler1988anomalies,snowberg2010favorite}: low-probability
contracts resolve true less often than their price implies.

How much less is a question this sample cannot settle, and the way one is tempted
to answer it is the mistake this paper is about. The largest deviation ---
contracts in the 0.20--0.30 band resolving true \CalWorstActual{} of the time
against a market-implied \CalWorstMarket{}, $n = \CalWorstN{}$ --- is the
\emph{minimum} over \CalNBins{} bins, and under the null that the market is
calibrated its standard error is \CalWorstSE{}, putting the deviation at
\CalWorstZ{}\,SE before any correction for having looked at \CalNBins{} bins.
Quoting it alone would be selecting a maximum and reporting it as a measurement.

The claim that survives is the pooled one. Across the \CalBelowBins{} bins at or
below 0.55, covering \CalBelowN{} markets, a calibrated market implies
\CalBelowExp{} resolutions in favour; \CalBelowObs{} occurred. That is
$z = \CalBelowZ{}$, $p = \CalBelowP{}$ two-sided --- the right sign, the right
order of magnitude, and not significant at conventional levels. The curve is also
not monotone: \CalWrongSign{} of the \CalNBins{} bins deviate in the opposite
direction, including the lowest bin.

The honest summary is that \CalN{} markets are consistent with the textbook bias
and too few to establish it.

\subsection{What the data does not support}

The forecasting system built on top of that calibration did not beat the market.
On a backtest of \BtN{} resolved events using prices from \BtDays{} days before
resolution, its mean Brier score \citep{brier1950verification} was \BtBrier{}
against a market-consensus \BtMarketBrier{}, and it produced a better forecast
than the market on \BtBeat{}\% of events. A separate zero-shot evaluation over
\ZsN{} events gave \ZsBrier{} against \ZsMarketBrier{} for the market, beating it
on \ZsBeat{}\% of events --- close to parity, on a sample far too small to
distinguish from it.

Both samples are small and neither supports a claim in either direction with any
confidence. We report them because the project's public materials at one point
claimed a 70\% beat rate over roughly two thousand predictions, and we could not
reproduce that figure: the per-prediction records behind it are not in the
repository or its history, and the evaluation artifacts that do survive are the
two above. Absent the underlying data, the claim should be treated as
unsupported.

\subsection{Why this belongs here}

The market case is the short-budget selection problem in a setting where the
consequence is money rather than a leaderboard row. A forecaster that picks,
among many candidate strategies, the one that scored best on the events it was
evaluated against, will report an edge it does not have --- and prediction markets
have small effective sample sizes, so the noise floor is high and the illusion is
correspondingly easy to produce. The curve in Figure~\ref{fig:calibration} is worth what it is worth precisely
because it is not the output of a search: it is one fixed estimator applied once,
and it is reported with the sample size that limits it.

\section{Recommendations}
\label{sec:reco}

The following is not a research agenda. It is the list of checks that would have
caught what we missed, ordered by how cheap they are.

\begin{enumerate}[leftmargin=1.8em]

\item \textbf{Report realised wall-clock, not the budget you passed.} One extra
column. A framework that overruns its budget by 2$\times$ is not being compared
at that budget, and the gap is invisible in every score-based statistic. If
median realised time differs across frameworks by more than a few percent, the
comparison is between different amounts of compute.

\item \textbf{Report how many candidates the search evaluated.} Together with the
test-set size this is enough for a reader to compute
$\sigma\sqrt{2\ln K}$ and decide whether your margins are resolvable. We did not
log $K$ in the original runs, which is why \S\ref{sec:confound-selection} has to
estimate it.

\item \textbf{Trace the evaluation data through the search code.} The defect here
is one variable: the dictionary passed to the scorer contained the test split.
The question to ask of any search loop is not ``do we hold out a test set'' but
``how many times is a number computed from it read by code that makes a
decision''. The answer must be one.

\item \textbf{Compare your margins against the noise floor before believing
them.} On a \TestSizeMed{}-row test set, accuracy differences below roughly one
point are not resolvable by a single split, whatever the $p$-value of an
aggregate test over datasets says. Our own sign test over \NData{} datasets
returns $p = \HOrcFlP{}$ for an effect we then show is largely selection bias:
the bias is present on every dataset, and the test asks only whether the sign is
consistent, not whether the effect is real.

\item \textbf{Prefer paired, same-run measurements when auditing a protocol.}
Re-running under a corrected protocol and comparing aggregate win rates confounds
the protocol change with run-to-run variance. Recording both estimands inside a
single run --- as in \S\ref{sec:corrected}, at the cost of roughly one extra
prediction per candidate --- isolates the protocol.

\item \textbf{Enforce budgets externally.} A deadline check between iterations is
not a time limit. Run the search in a killable process, or wrap each candidate in
a timeout. If the harness runs several datasets concurrently, either pin cores or
accept that the budget is meaningless.

\item \textbf{Treat a ``tie'' threshold as a claim about precision.} Declaring
ties at $10^{-6}$ on a metric whose standard error is $10^{-2}$ means near-ties
are resolved by noise. If the estimator on one side is a maximum over many
draws, they are resolved in its favour.

\end{enumerate}

None of these requires a containerised harness or a benchmark suite. The first
two are logging. The rest are ten lines of code and a habit. The question they
serve --- whether a more elaborate method actually beats a well-tuned simple
baseline once the comparison is set up honestly --- is not specific to AutoML
\citep{zhang2026calibrated}, and neither are the ways of getting it wrong.

\section{Limitations}
\label{sec:limitations}

\paragraph{One split per dataset.} Each dataset contributes a single 80/20
holdout comparison, where AMLB \citep{gijsbers2024amlb} would use 10-fold
cross-validation. Aggregating over \NData{} datasets stabilises the win rate but
does nothing for per-dataset variance --- and per-dataset variance is precisely
the quantity that the selection defect exploits. A cross-validated harness would shrink the
per-dataset noise --- by roughly $\sqrt{5}$ in effective sample size for 10-fold
against an 80/20 holdout, less once fold correlation is accounted for --- and
with it the effect we measure. Our
numbers therefore describe the regime that informal comparisons actually run in,
not the regime a careful benchmark would use.

\paragraph{The dataset collection is not a curated suite.} We filtered OpenML by
size and quality rather than adopting a fixed benchmark suite, and \DupNames{}
names recur because OpenML versions the same underlying data. \S\ref{sec:naive} reports the headline both ways; \orc{}'s share is
essentially unchanged, though the split between the other two moves by a couple
of points.
Aggregate percentages should still be read as descriptive of this collection
rather than of tabular ML in general.

\paragraph{\ag{} is handicapped.} It ran at \texttt{medium\_quality} without the
optional \texttt{fastai} dependency, so one model in its portfolio was skipped on
every dataset. Its numbers are a lower bound, which cuts against our own headline
in the direction of making \orc{} look better than it is.

\paragraph{The corrected run is a subsample.} Re-running all \NData{} datasets
three ways at a hard 60-second budget takes on the order of a day of wall-clock
on this machine; we re-ran a stratified random subsample instead. The paired
per-dataset gap (Figure~\ref{fig:oraclegap}) is well determined at this sample
size; the corrected aggregate win rates carry a standard error of a few
percentage points and should not be read to one decimal place.

\paragraph{The order-statistics estimate is an upper bound.} It assumes
independent Gaussian errors across candidates. Real candidates are correlated ---
same test rows, overlapping model families --- so the effective number of draws is
below $K$ and the true inflation is smaller than $\sigma\sqrt{2\ln K}$. We use it
to establish scale, not to predict a number.

\paragraph{The validation split costs \orc{} data.} Under the corrected protocol
the search fits on 64\% of each dataset rather than 80\%. Refitting the selected
model on the full training split before the final evaluation removes most of this
disadvantage but not the part that affects which model gets selected on small
datasets.

\paragraph{One system, one platform.} We measure the bias produced by one
search loop on one ARM64 machine. The mechanism is general and the magnitude is
computable from $K$ and test-set size, but the specific percentages are not
portable.

\section{Conclusion}

A 1{,}661-line random search over a fixed scikit-learn model pool appeared to beat
\flaml{} and \ag{} on \NData{} OpenML datasets, by margins that were consistent
across task types, stable across two budgets, and significant at
$p = \HOrcFlP{}$. It did so because it reported the best of dozens of test-set
evaluations while its competitors reported one, and because it took a median
\RatioOrcAgMed{}$\times$ the per-dataset wall-clock \ag{} did under the same
nominal budget.

Correcting the protocol takes it from \FairOrigOrc{}\% to \FairOrc{}\% on the
re-run subset, where no pairwise difference against either competitor is
significant.

We have tried to leave behind something more useful than a retraction. The
paired-measurement design in \S\ref{sec:corrected} --- recording both estimands
inside one run, at the cost of one prediction per candidate --- turns a protocol
audit into a controlled experiment, and it is what let us say that the selection
rule was worth \SelRulePts{} points and compute the rest. It also corrected us:
the closed-form screen we started from, $\sigma\sqrt{2\ln K}$ from the test-set
size and the candidate count, overstated the realised bias by about five times,
because candidates scored on shared rows cancel most of the noise. The screen is
still worth running --- it is two numbers every harness already logs, and a margin
beneath it cannot be trusted --- but it bounds rather than predicts, and we would
not have known by how much without measuring.

The broader point is about where evaluation error concentrates. Careful AutoML
benchmarks run at hour-scale budgets with cross-validated folds, where the noise
floor is low and a one-point margin means something. The comparisons that
circulate run at second-scale budgets on single splits, where the noise floor is
high, the number of selection events is large, and the two are related by the
budget.

\bibliographystyle{plainnat}
\bibliography{references}

\appendix
\section{Model pool and search space}
\label{app:pool}

\begingroup\sloppy
\paragraph{Baseline pool (fitted first, always).}
The four regression entries marked $\dagger$ and the post-search ensemble were
added after the reported runs (\S\ref{sec:setup}).
Regression: \texttt{LinearRegression}, \texttt{Ridge}$(\alpha{=}1)$,
\texttt{RandomForest}$(100)$, \texttt{ExtraTrees}$(100)$,
\texttt{GradientBoosting}$(100)$, \texttt{HistGradientBoosting}$(200)$,
\texttt{XGBoost}$(100)$, \texttt{LightGBM}$(100)$,
\texttt{ElasticNet}$^\dagger(\alpha{=}1, \ell_1{=}0.5)$,
\texttt{KNN}$^\dagger(k{=}5)$, \texttt{SVR}$^\dagger(C{=}1)$,
\texttt{CatBoost}$^\dagger(200)$.
Classification: \texttt{LogisticRegression}, \texttt{RandomForest}$(100)$,
\texttt{ExtraTrees}$(100)$, \texttt{GradientBoosting}$(100)$,
\texttt{HistGradientBoosting}$(200)$, \texttt{XGBoost}$(100)$,
\texttt{LightGBM}$(100)$. Linear, KNN and SVM baselines are fitted on
standardised features; tree models on raw features.

\endgroup

\paragraph{Search space.} Proposals draw uniformly from discrete grids:
$n_{\text{estimators}} \in \{100, 200, 300, 500, 800\}$,
learning rate $\in \{0.01, 0.03, 0.05, 0.1, 0.2\}$,
max depth $\in \{3,4,5,6,7,8,10,15,20,\text{None}\}$ (and $-1$ for LightGBM),
subsample and column-sample $\in \{0.6,\dots,1.0\}$,
$\text{num\_leaves} \in \{15, 31, 50, 80, 127\}$,
Ridge/Lasso $\alpha \in \{0.001,\dots,100\}$,
$k \in \{3, 5, 7, 10, 15\}$ with uniform or distance weights,
$C \in \{0.1, 1, 10\}$, logistic $C \in \{0.01, 0.1, 1, 10\}$,
$\texttt{min\_samples\_split} \in \{2, 5, 10\}$,
$\texttt{min\_samples\_leaf} \in \{10, 20, 30, 50\}$,
AdaBoost $n \in \{50, 100, 200\}$, ElasticNet
$\ell_1 \in \{0.1, 0.3, 0.5, 0.7, 0.9\}$, and XGBoost
$\alpha, \lambda \in \{0, 0.01, 0.1, 1\}$. The classification grids are
slightly coarser than the regression grids.
Preprocessors, proposed only for linear, KNN and SVM families:
standard, min-max, robust, quantile, and Yeo--Johnson power transforms.

\paragraph{Proposer schedule.} Iterations 1--5 take one proposal from each
untried tree family, in a fixed order that leaves LightGBM unforced when six
families are installed. From iteration 6, the currently best family is re-proposed
with probability $\min(0.5 + 0.01t, 0.75)$; otherwise a family is drawn 70/20/10
from tree, linear, and other. From iteration 11, each proposal is with
probability $0.15$ a fixed voting ensemble instead (tree families for
regression; random forest plus histogram gradient boosting for classification). Proposals are hashed by their
description and duplicates are discarded without evaluation.

\paragraph{Post-search ensemble.} The best three models of distinct classes that
used no preprocessor are combined by soft voting, weighted by $1/\text{score}$
for minimised metrics and by score for maximised metrics. It is kept only if it
beats the best single model.

\section{Reproduction}
\label{app:repro}

All result files are JSONL with one record per dataset, containing the dataset
ID, task type, shape, each framework's score, selected model and wall-clock, and
the decided winner.

\begin{itemize}
\item \texttt{experiments/results/flaml\_strict30s\_*.jsonl} --- 30-second
two-way run (\S\ref{sec:naive}), serial execution.
\item \texttt{experiments/results/multi\_framework\_*.jsonl} --- 60-second
three-way run, four workers. The file suffixed
\texttt{regression\_only\_orcetra\_improved} is a later re-run of the regression
tasks under a different machine load; it is analysed in \S\ref{sec:splice} and
deliberately \emph{not} merged into the headline.
\item \texttt{paper/figures/data/fair160.jsonl} --- protocol-corrected run
(\S\ref{sec:corrected}), produced by \texttt{experiments/fair\_benchmark.py}.
Each record carries the three estimands $A$, $B$ and $C$, the validation score,
the per-candidate trace of (validation, test) pairs, the number of candidates
evaluated, the search and refit times, and whether the search was killed at the
deadline.
\end{itemize}

To regenerate the paper's numbers and figures from those files:

\begin{quote}\ttfamily\small
python3 paper/figures/make\_numbers.py\\
python3 paper/figures/make\_figures.py\\
latexmk -pdf paper/main.tex
\end{quote}

\noindent
\texttt{make\_numbers.py} writes \texttt{paper/numbers.tex}, a file of LaTeX
macros; the prose contains no hard-coded quantities, so a changed result file
propagates into the text on the next build.

\section{Reconciling the paper with the repository}
\label{app:readme}

The project README reports the 60-second three-way result as
\WinOrcAll{}\,/\,\WinAgAll{}\,/\,\WinFlAll{}\,/\,\WinTieAll{}\%, which is what
Table~\ref{tab:main} reports, over the same \NData{} datasets of the single
consistent sweep. The two agree.

They agree only because the README predates the regression re-run.
Deduplicating every \texttt{multi\_framework\_*.jsonl} by dataset ID instead
gives \MergedOrc{}\,/\,\MergedAg{}\,/\,\MergedFl{}\,/\,\MergedTie{}\% over
\MergedN{} rows; \S\ref{sec:splice} shows why that figure should not be used.
Anyone reproducing our numbers should exclude
\texttt{multi\_framework\_20260331\_1917\_regression\_only\_orcetra\_improved.jsonl}
from the headline, as \texttt{make\_numbers.py} does. All of these are in any
case numbers produced by the protocol this paper argues is invalid.

\end{document}